\documentclass[conference,a4paper]{IEEEtran}
\IEEEoverridecommandlockouts
\usepackage{cite}
\usepackage{amsmath,amssymb,amsfonts}
\usepackage{algorithmic}
\usepackage{graphicx}
\usepackage{textcomp}
\usepackage{xcolor}
\usepackage{academicons}
\usepackage{url}
\usepackage{hyperref}
\usepackage{gensymb}
\usepackage{caption}
\usepackage[T1]{fontenc}

\def\BibTeX{{\rm B\kern-.05em{\sc i\kern-.025em b}\kern-.08em
    T\kern-.1667em\lower.7ex\hbox{E}\kern-.125emX}}
\begin{document}

\newcommand{\orcid}[1]{\href{https://orcid.org/#1}{\includegraphics[scale=0.07]{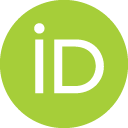}}}

\title{Catadioptric Stereo on a Smartphone
}
\author{\IEEEauthorblockN{Kristijan Bartol\textsuperscript{\orcid{0000-0003-2806-5140}}, David Bojani\'{c}\textsuperscript{\orcid{0000-0002-2400-0625}}, Tomislav Petkovi\'{c}\textsuperscript{\orcid{0000-0002-3054-002X}}, Tomislav Pribani\'{c}\textsuperscript{\orcid{0000-0002-5415-3630}}}
\IEEEauthorblockA{\textit{University of Zagreb} \\
\textit{Faculty of Electrical Engineering and Computing}\\
Department of Electronic Systems and Information Processing \\
name.surname@fer.hr}
}

\maketitle

\begin{abstract}

We present a 3D printed adapter with planar mirrors for stereo reconstruction using front and back smartphone camera. The adapter presents a practical and low-cost solution for enabling any smartphone to be used as a stereo camera, which is currently only possible using high-end phones with expensive 3D sensors. Using the prototype version of the adapter, we experiment with parameters like the angles between cameras and mirrors and the distance to each camera (the stereo baseline). We find the most convenient configuration and calibrate the stereo pair. Based on the presented preliminary analysis, we identify possible improvements in the current design. To demostrate the working prototype, we reconstruct a 3D human pose using 2D keypoint detections from the stereo pair and evaluate extracted body lengths. The result shows that the adapter can be used for anthropometric measurement of several body segments.
%For future work, we will build a more rigid adapter with fixed parameters and larger mirrors to allow a wider baseline for successful stereo reconstruction.
\end{abstract}

\begin{IEEEkeywords}
catadioptric stereo, planar mirrors, smartphone, adapter, camera calibration, 3D reconstruction, body measurement
\end{IEEEkeywords}

\section{Introduction}

Stereo vision is a well-known approach for 3D reconstruction. It is a popular as it only requires two cameras and imposes relatively few constraints, such as textured scene and reasonably-wide baseline, compared to other 3D reconstruction techniques \cite{anthropometry-review}. Stereo is important for numerous applications, such as 3D scanning \cite{anthropometry-review}, cultural heritage replications \cite{3d-scanning-cultural-heritage}, SLAM \cite{orb-slam2}, etc. 

Regarding smartphones, 3D reconstruction is becoming more and more accessible with embedded ToF sensors on Androids and iPhones, but these technologies are still not affordable for the mass. Alternatively, an attempt towards stereo was made on a smartphone equipped with multiple back cameras \cite{mas18}; however, the baseline between these cameras is very small, which degrades the reconstruction performance. Finally, there are works which take advantage of built-in video projectors \cite{3d-registration-direction-sensor, 3d-structured-light-on-smartphone}; still, most smartphones and tablets are not equipped with them. We therefore present a low-cost solution for stereo on a smartphone --- proposing a novel design of a 3D printed adapter with mirrors (catadioptric stereo) depicted in Fig. \ref{fig:adapter}, front and back camera, on standard smartphone into common field-of-view (FOV).

In general, catadioptric systems consist of mirrors and lenses \cite{rectified-catadioptric-stereo}. By using mirrors, a common part of the scene can be imaged from multiple views, which allows 3D reconstruction \cite{zisserman}. Most of the previously presented catadioptric systems use a single camera with multiple planar mirrors \cite{design-of-a-single-lens-stereo, a-stereo-viewer-single-camera, panoramic-stereo, stereo-with-mirrors, planar-catadioptric-stereo, stereovision-with-single-camera-multiple-mirrors}. Several works use prisms in combination with planar mirrors \cite{single-camera-stereo-using-prism-and-mirrors, biprism-stereo}. The remaining works analyze the use of hyperbolic \cite{omnidirectional-stereo-vision} and parabolic mirrors \cite{catadioptric-self-calibration, stereovision-with-single-camera-multiple-mirrors}.

\begin{figure*}[h!]
	\centering
	\includegraphics[width=.75\linewidth]{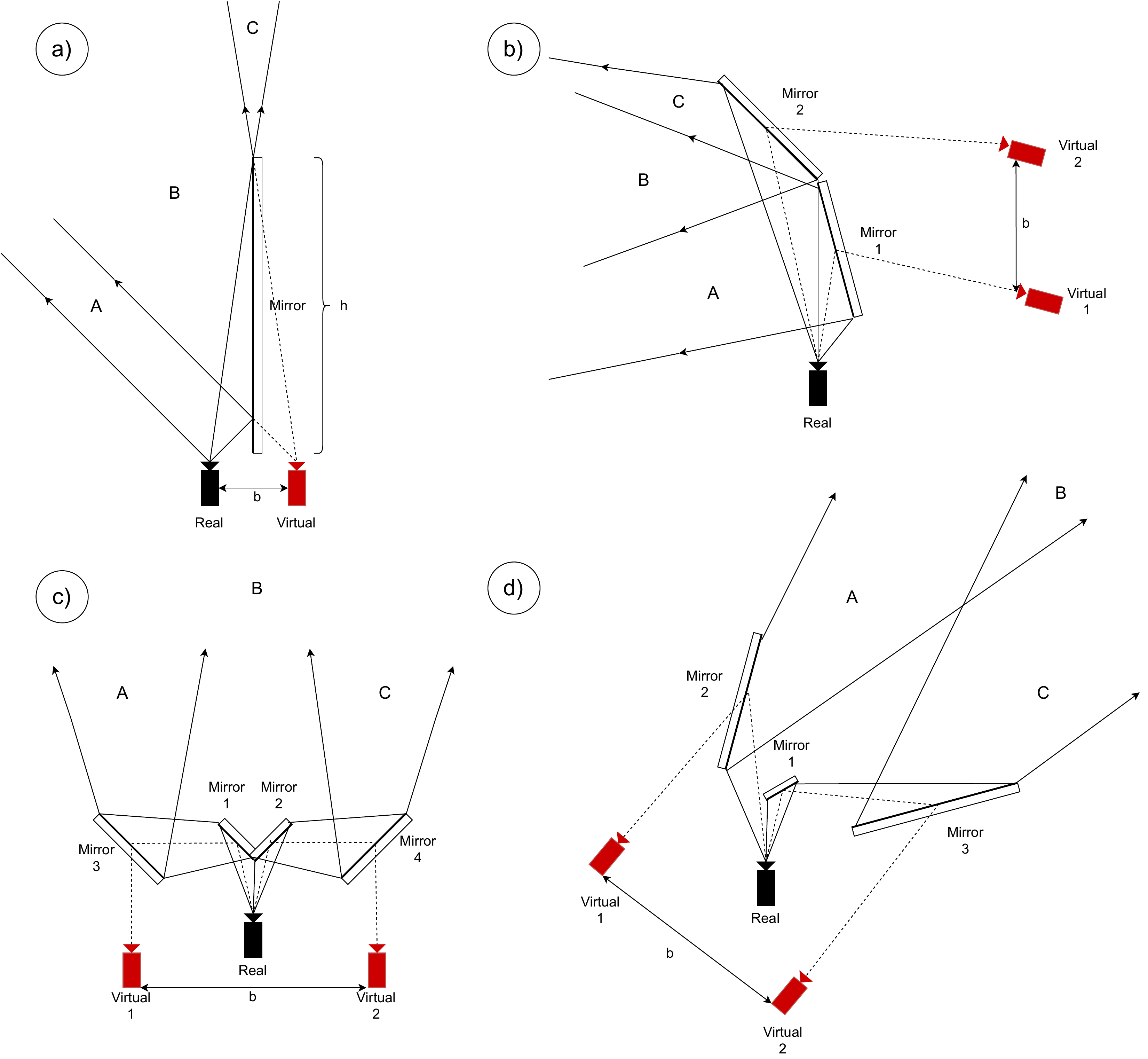}
	%\captionsetup{justification=centering}
	\caption{Four previously proposed planar-mirror catadioptric configurations. The Figure shows real (black) and corresponding virtual (red) cameras, mirror placement, baselines, and three areas, A, B, and C, for each setup. Letter B represents a common area, visible by the stereo pair, while A and C are unused areas.}
	\label{fig:single-camera-configurations}
\end{figure*}

Regarding planar mirrors, Fig. \ref{fig:single-camera-configurations} shows four different catadioptric stereo configurations, previously described by Gluckman and Nayar \cite{rectified-catadioptric-stereo}. All four systems use a single camera. The simplest configuration (Fig. \ref{fig:single-camera-configurations}a) creates a single virtual camera, using the reflection in the mirror. A stereo pair consists of a real and a virtual camera. By moving a camera from the mirror, the baseline increases, but the common FOV between the images proportionally decreases. A two-mirror configuration (Fig. \ref{fig:single-camera-configurations}b) produces two virtual cameras and these cameras comprise a stereo pair. The third configuration ((Fig. \ref{fig:single-camera-configurations}c) uses four mirrors, as shown in a lower part of Fig. \ref{fig:single-camera-configurations}. The advantage of third configuration over the second is that symmetric mirror setup produces virtual cameras without relative rotation, which is more suitable for stereo reconstruction, i.e. it does not require rectification \cite{zisserman}. However, the third configuration is not practical for 3D reconstruction of nearby objects, as the common FOV is relatively far from the virtual cameras. The fourth configuration (Fig. \ref{fig:single-camera-configurations}d) fixes the above issue. Our system uses two planar mirrors and two cameras.

To the best of our knowledge, no peer-reviewed work\footnote{Note that there exist a webpage where multi-camera catadioptric systems are described \cite{catadioptric-website}.} has been dedicated to extending multi-camera systems with mirrors or prisms to enable stereo on a standard smartphone. The analysis of such catadioptric systems makes sense today, with the advent of high-resolution front-back smartphone camera configurations. 

In the remainder of the paper, we describe previous work on using mirrors for enabling stereo, then we present our adapter and its parameters. In the experimental section, we analyze the effect of the parameters to the FOV, calibration procedure, and reconstruction results.

\section{Related Work}

We describe related work and comment on catadioptric stereo configurations of interest in more detail.

One of the first attempts to create catadioptric stereo using planar mirrors is by Goshtasby and Gruver in 1993 \cite{design-of-a-single-lens-stereo}. They propose a simple two-mirror configuration and use mirror reflections to produce two virtual cameras. Inaba et al. \cite{a-stereo-viewer-single-camera} present a four-mirror adapter without adjustable mirrors, similar to the third configuration in Fig. \ref{fig:single-camera-configurations}. Gluckman and Nayar \cite{catadioptric-stereo-using-planar-mirrors, planar-catadioptric-stereo, rectified-catadioptric-stereo} analyze the geometry and calibration of catadioptric stereo rig. They also describe several planar-mirror configurations, some of them shown in Fig. \ref{fig:single-camera-configurations}, to achieve stereo system that do not require rectification for each processed frame.

Several works used the combination of (bi-)prisms and mirrors to obtain stereo configuration from a single camera. Lee et al. \cite{biprism-stereo} used a biprism placed more than 10\,cm in front of the camera to obtain images from two virtual cameras. The idea was extended and generalized using triprism in \cite{prism-based-single-lens-stereo}. Two works, \cite{single-camera-system-captures-3d-images} and \cite{single-camera-stereo-orientable-optical-axes}, used a prism and two mirrors, where the prism is used as a reflective surface, similar to the third setup in Fig. \ref{fig:single-camera-configurations}.

Regarding curved mirrors, Southwell et al. (1993) \cite{panoramic-stereo} produce a panoramic stereo setup using a single convex mirror. Baker and Nayar (1998) \cite{a-theory-of-catadioptric-image-formation} showed that, among curved mirrors, only quadric mirrors (parabolic, elliptic, and hyperbolic), can be placed in a configuration with a pinhole camera, such as ours, where all the rays reflected from the mirror intersect at a single viewpoint. Kang \cite{catadioptric-self-calibration} proposed a self-calibrating setup using camera and a parabolic mirror. Mouaddib et al. \cite{stereovision-with-single-camera-multiple-mirrors} analyzes several catadioptric configurations, including multiple planar mirror and hyperbolic mirrors, propose criterions and compare the presented configurations.

In contrast to prior work, we analyze stereo configuration using front and back smartphone camera and two planar mirrors.

\section{Catadioptric Stereo Adapter}
\label{sec:catadioptric-stereo-adapter}

\begin{figure}[h!]
	\centering
	\includegraphics[width=1.\linewidth]{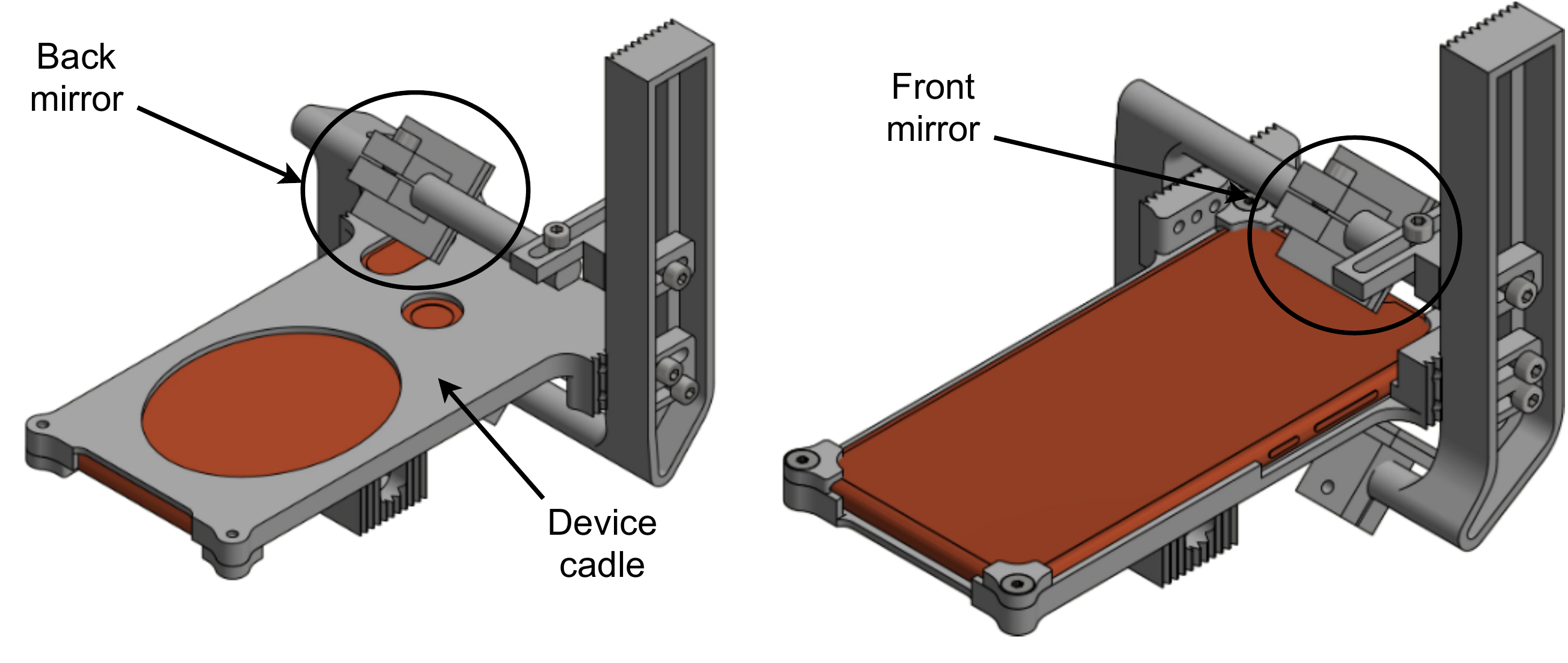}
	\caption{The adapter design in a regular 3D view. The Figure features front and back mirrors and a smartphone mask, to keep the device fixed. The smartphone is almost normally usable for image recording, as the buttons on the side are available.}
	\label{fig:design-3d}
\end{figure}

\begin{figure}[h!]
	\centering
	\includegraphics[width=1.\linewidth]{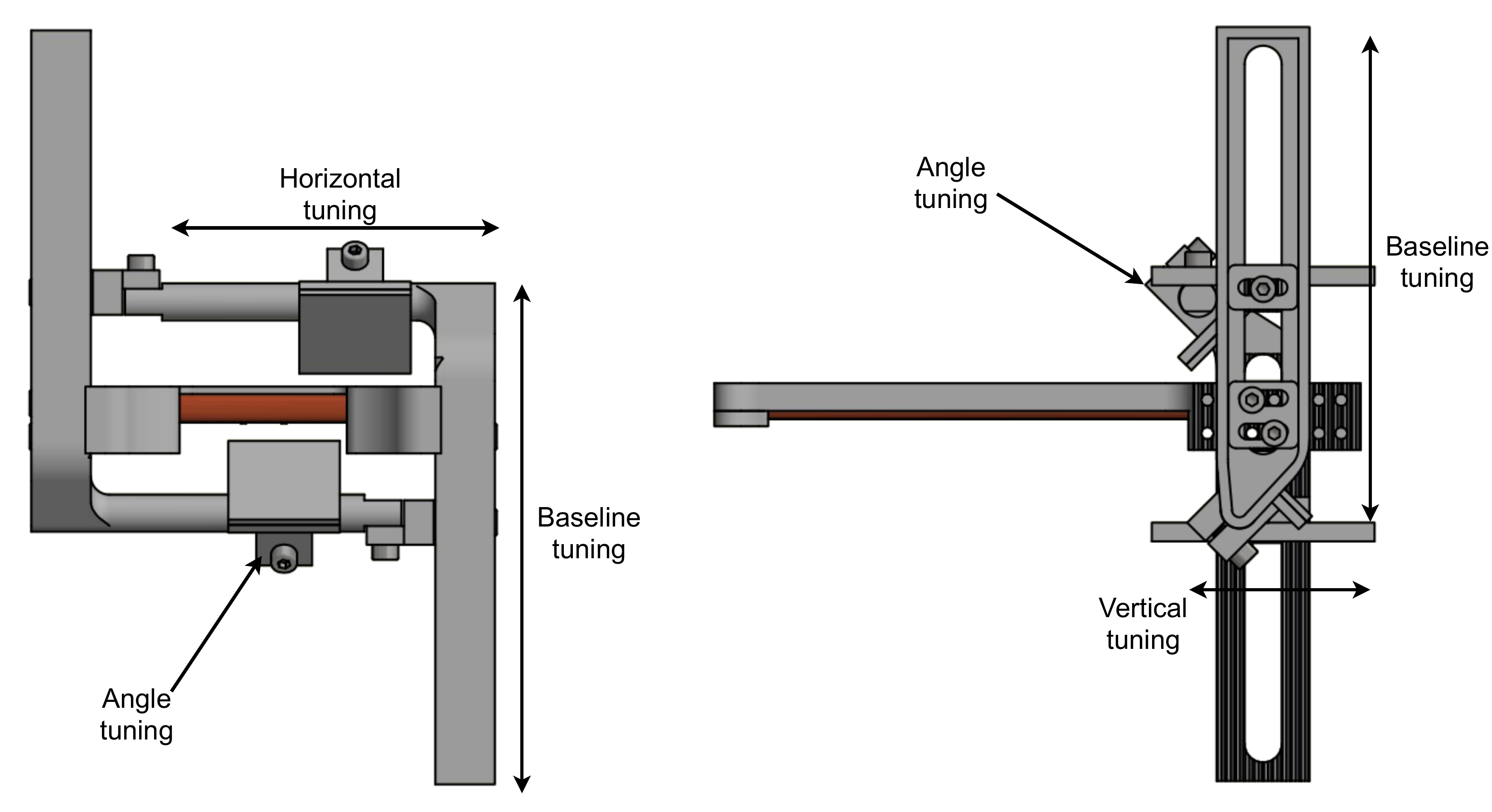}
	\caption{The adapter design in forward and side view. The Figure features the parameters for adjusting mirror position (vertical, horizontal, and angular).}
	\label{fig:design-flat}
\end{figure}

\begin{figure}[h!]
	\centering
	\includegraphics[width=.75\linewidth]{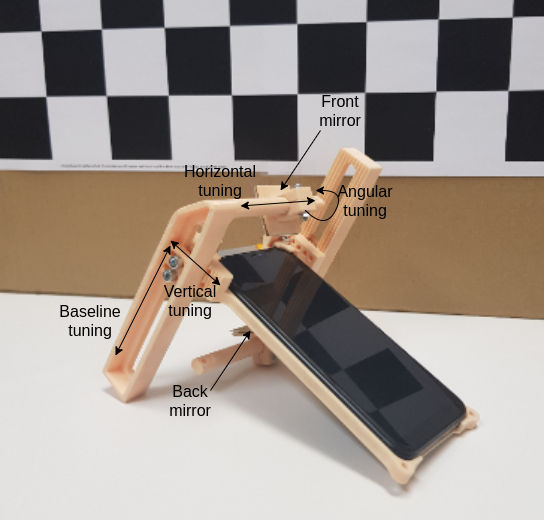}
	\caption{The 3D printed adapter, pointing out front and back mirrors and baseline tuning.}
	\label{fig:adapter}
\end{figure}

The 3D printed adapter consists of two mirrors to enable two cameras to record a common part of the scene (see Fig. \ref{fig:adapter}). To accomodate for different camera placements and to be able to experiment with different baselines and FOVs, we have several degrees of freedom:

\begin{itemize}
    \item baseline (Fig. \ref{fig:design-flat}),
    \item mirror angle (Fig. \ref{fig:design-3d} and \ref{fig:design-flat}),
    \item vertical and horizontal tuning (Fig. \ref{fig:design-flat}).
\end{itemize}

%Horizontal and angular tuning are done by unscrewing the screws behind the mirrors, and similarly for vertical placement. 
Horizontal and vertical tuning allow moving the mirrors to compensate for various camera positions on different smartphones. Baseline and angular tuning affect the common FOV of the virtual cameras (see the remainder of the section). The current version of phone cadle is designed specifically for Xiaomi Mi A2 smartphone, based on its dimensions, to perfectly sit into the mask and stay fixed during reconstruction for multiple experiments.

\subsection{Baseline and Angular Tuning}

%Compared to Fig. \ref{fig:single-camera-configurations} and prior work, we show our camera-mirror-smartphone setup in Fig. \ref{fig:smartphone-stereo-mirrors}. 
When building catadioptric stereo, some parts of the images are unusable, on two levels. First, the mirror of each individual camera does not cover the whole FOV, and, second, the common FOV of the virtual cameras is smaller than the original, single-camera FOV. 
%Note that one of the advantages of smartphone cameras is their wide angle. Our Xiaomi testing device has 80\textdegree{} FOV in both front and back camera. 
In this Section, we analyze how the baseline, angles, and mirror sizes affect the common FOV of the virtual cameras. 
%In the remainder of the section, we analyze how baseline and angular tuning affect the common FOV of the virtual cameras. 
More specifically, we derive two quantities: 

\begin{itemize}
    \item the percentage of the common FOV, $\%_{\text{FOV}}$, retained compared to the original, single-camera FOV, and
    \item the minimal distance between the virtual camera and the person, $d_{\text{min}}$, needed to fit the average person ($h_{\text{avg}}=1.8\,$m) into the common FOV.
\end{itemize}

\begin{figure}[t!]
	\centering
	\includegraphics[width=1.\linewidth]{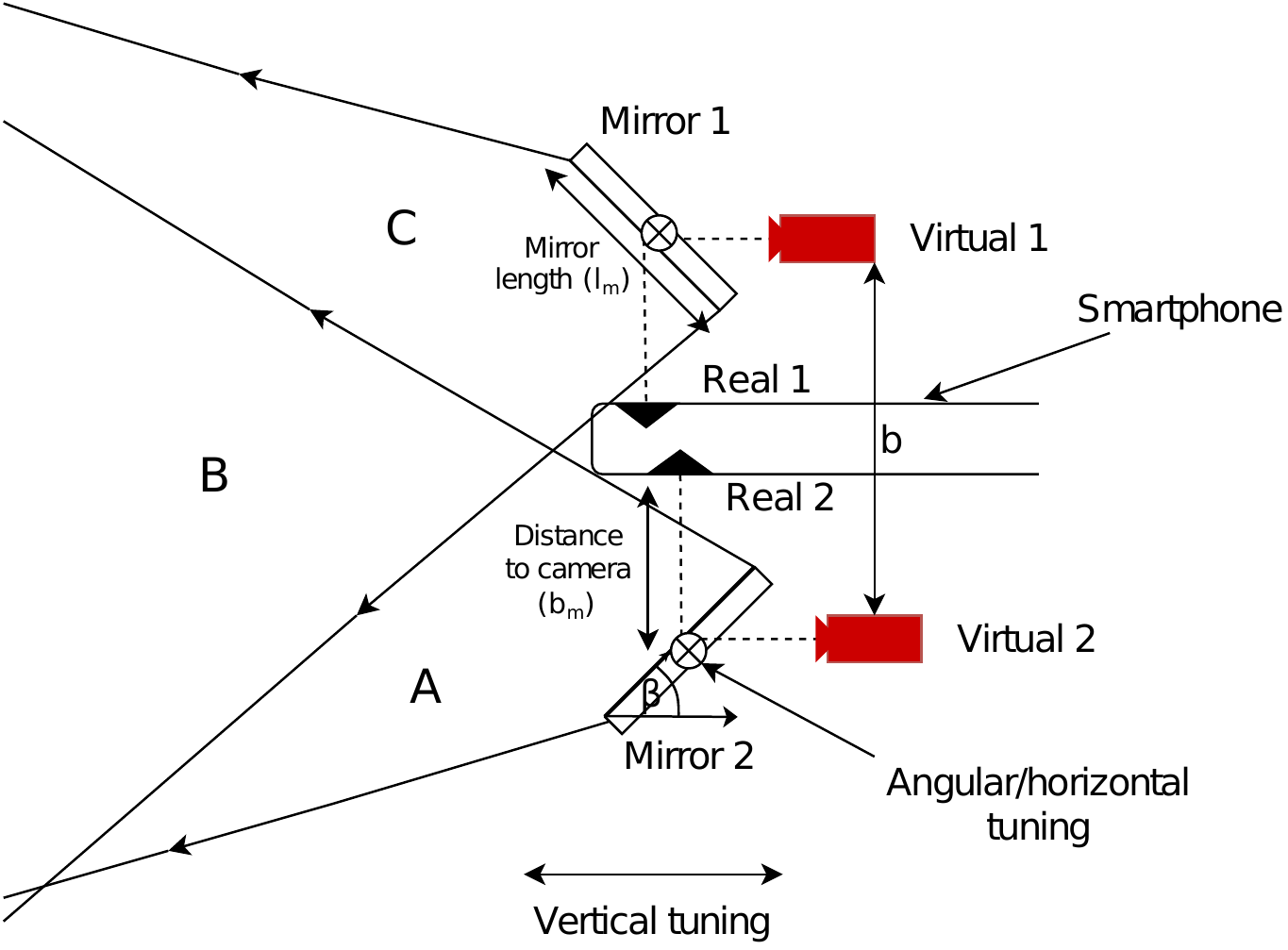}
	\caption{The designed catadioptric stereo using front and back smartphone camera. Letter B shows the common FOV of the two virtual cameras. To adjust for the desired stereo pair properties and for different real camera positions, the mirrors can be moved in all three axes.}
	\label{fig:smartphone-stereo-mirrors}
\end{figure}

First, we derive the angle $\alpha_{\text{virtual}}$ for a single camera (Fig. \ref{fig:single-camera-analysis}). Then, we use the angle $\alpha_{\text{virtual}}$ to calculate the minimal distance, $d_{\text{min}}$, needed to record a person, and the retained common FOV, $\%_{\text{FOV}}$. We take into account the following parameters (see Fig. \ref{fig:smartphone-stereo-mirrors}):

\begin{itemize}
    \item distance $b_m$ between the mirror and the camera,
    \item mirror length $l_m$,
    \item angle of the mirror $\beta$.
\end{itemize}

Note that changing the distance $b_m$ directly affects the baseline $b$ (see Fig \ref{fig:smartphone-stereo-mirrors}). Mirror length $l_m$, for squared mirrors such ours, results in a $P = l_m^2\,$cm surface. 
%The goal of the following analysis is to determine the percentage of FOV that is acquired, based on several previously mentioned parameters:

\textbf{Individual FOV analysis.} The FOV of the virtual camera, $\alpha_{\text{virtual}}$, can be derived as a sum of left and right angles, $\alpha_{\text{virtual}} = \alpha_L + \alpha_R$, as shown in Fig. \ref{fig:single-camera-analysis}. The left angle, $\alpha_L$, spans between the middle of mirror and its left end, i.e. right end in case of the right angle, $\alpha_R$. The angles can be calculated using the tangent function, as follows:

\begin{equation}
\label{eq:alpha_L}
    \tan{\alpha_L} = \frac{l_{m\_\text{proj}}}
    {2b_{m\_\text{down}}}
\end{equation}

\begin{equation}
\label{eq:alpha_R}
    \tan{\alpha_R} = \frac{l_{m\_\text{proj}}}
    {2b_{m\_\text{up}}}
\end{equation}

The $l_{m\_\text{proj}}$ is the mirror length projected on the line perpendicular to the optical axis of the real camera. The values $b_{m\_\text{up}}$ and $b_{m\_\text{down}}$ are the distances of the upper and the lower corner of the mirror. The mirror projection length is simply $l_{m\_\text{proj}} = l_m \cos{\beta}$, and the distances are $b_{m\_\text{up}} = b_m - \frac{h_m}{2}$ and $b_{m\_\text{down}} = b_m + \frac{h_m}{2}$. Mirror height $h_m$ can be calculated as $h_m = l_m \sin{\beta}$. Virtual camera angle is therefore:

\begin{figure}[t!]
	\centering
	\includegraphics[width=1.\linewidth]{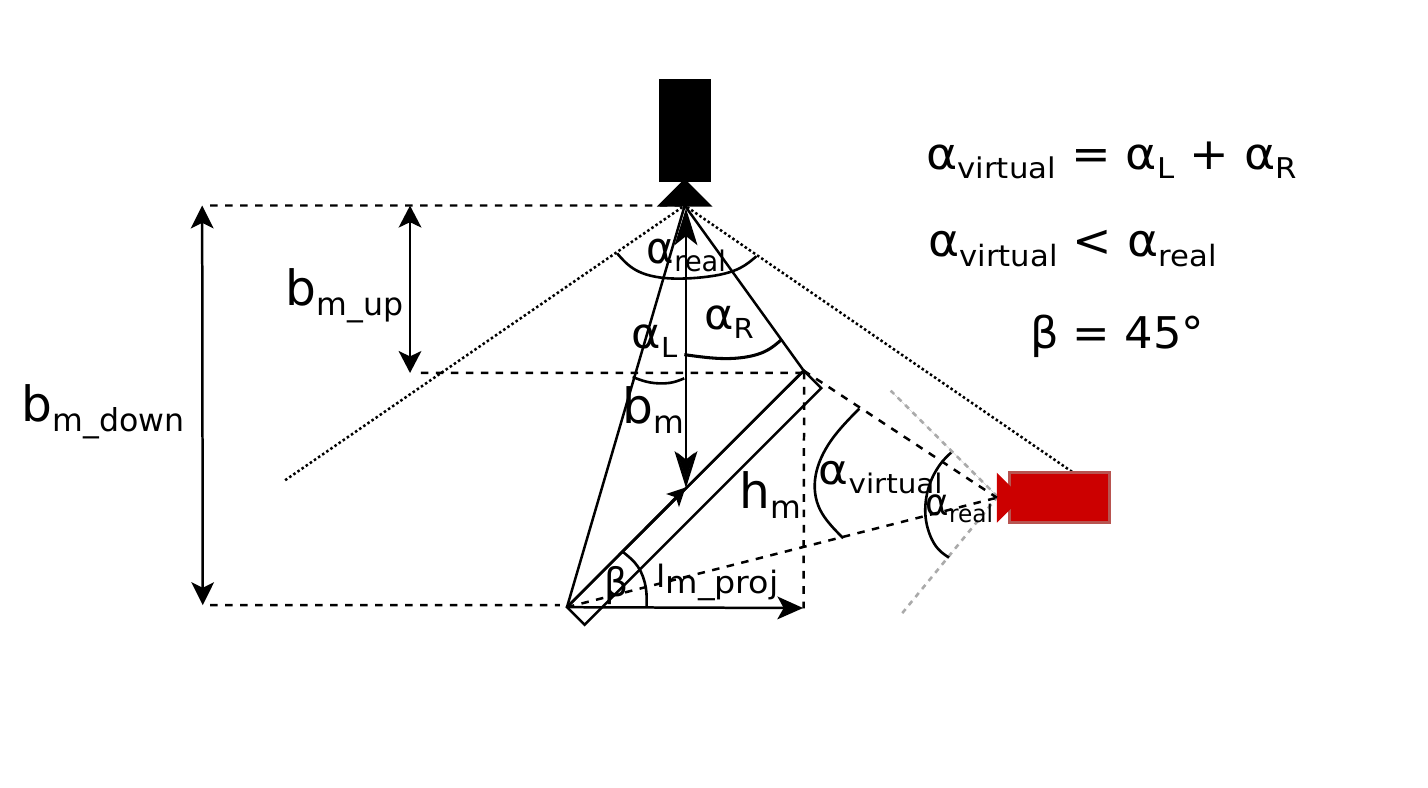}
	\caption{The analysis of the virtual camera FOV with respect to the distance between the real camera and the mirror, $d_m$, mirror length $l_m$, and mirror angle $\beta$. Note that $\beta=45$\textdegree{} is shown for simplicity, so that the optical axis of the virtual camera is perpendicular to the real.}
	\label{fig:single-camera-analysis}
\end{figure}

\begin{equation}
    \alpha_{\text{virtual}} = \tan^{-1} \left( \frac{l_m \cos{\beta}}{2 b_m + l_m \sin{\beta}}\right) + \tan^{-1} \left( \frac{l_m \cos{\beta}}{2 b_m - l_m \sin{\beta}} \right)
\end{equation}

The percentage of the retained FOV for an individual camera is $\%_{\text{FOV}} = \frac{\alpha_{\text{virtual}}}{\alpha_{\text{real}}}$.

\begin{figure}[t!]
	\centering
	\includegraphics[width=1.\linewidth]{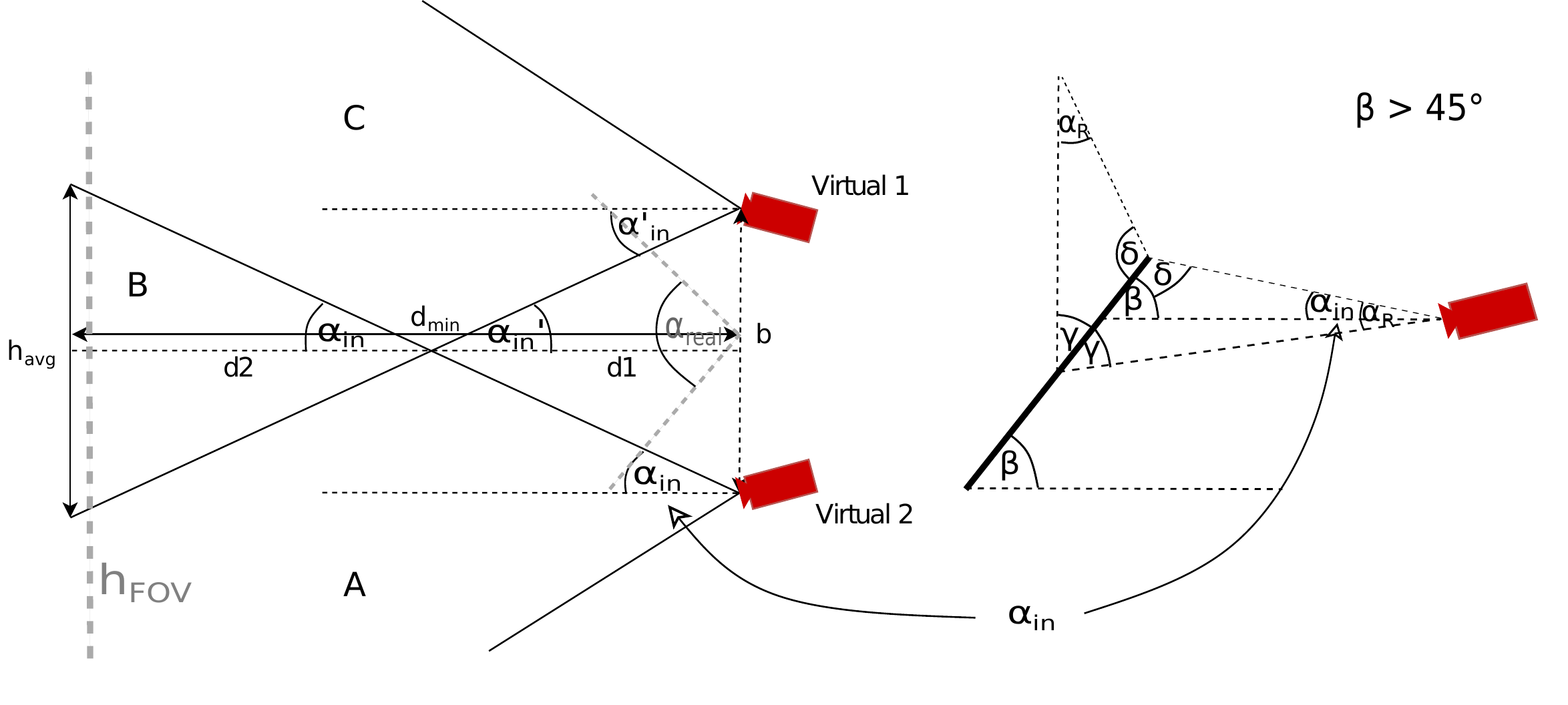}
	\caption{The analysis of virtual cameras and their common FOV, assuming mirror angle to be $\beta > 45\degree$.}
	\label{fig:two-camera-analysis}
\end{figure}

\textbf{Common FOV analysis.} In the second part, we analyze what happens when mirror angle $\beta > 45$\textdegree{} with respect to the common FOV (similar analysis can be done for $\beta < 45\degree$). We also find the minimal distance needed to record an average-height person. For simplicity, we assume that mirror angles $\beta$ for both mirrors are the same (therefore, $\alpha_{in} = \alpha_{in}'$, as seen in Fig. \ref{fig:two-camera-analysis}). We also assume that the real cameras are one beneath the other with respect to the smartphone, even though the physical cameras are not. The latter assumption does not significantly change the analysis, as the distances between the cameras are relatively small compared to the distance to the object, as will be shown in the remainder of the section.

Fig. \ref{fig:two-camera-analysis} shows a virtual camera configuration. The angle $\alpha_{in}$ can be calculated as $\alpha_{in} = 180\degree - \beta - \delta$. (see Fig. \ref{fig:two-camera-analysis}, right). The unknown angle is $\delta = 180\degree - \alpha_R - \gamma$, where $\gamma = 90\degree - \beta$. Finally, the inner angle is $\alpha_{in} = 90\degree + \alpha_R - 2\beta$.

The minimal distance between the virtual camera and the person is: 

\begin{equation}
    d_{\text{min}} = d_1 + d_2 = \frac{b + h_{\text{avg}}}
    {2\tan{\alpha_{in}}}.
\end{equation}

The height visible from the single, real camera can be calculated as $h_{\text{FOV}} = 2 d_{\text{min}} \tan{\frac{\alpha_{\text{real}}}{2}}$. Assuming that original single-camera FOV is $\alpha_{\text{real}} = 80\degree$ and $h_{\text{avg}} = 1.8$\,m, then the percentage of the retained, common FOV between the virtual cameras is:

\begin{equation}
    \%_{\text{FOV}} = \frac{h_{\text{avg}}}{h_{\text{FOV}}} =
    \frac{h_{\text{avg}}}
    {2d_{\text{min}}\tan{\frac{\alpha_{\text{real}}}{2}}} = \frac{1.8 \cdot 100}
    {2d_{\text{min}}\tan{40\degree}} \% .
\end{equation}

\textbf{Selecting parameters.} Based on the analysis and our empirical observations, we decided to set both mirrors to (roughly) $\beta=55$\textdegree{}, $b_m=2.5$\,cm apart from the device, and use a 3\,cm x 3\,cm ($l_m=3\,$cm) mirror surfaces. Based on the equations in the previous section, each virtual camera FOV is reduced to $\alpha_{\text{virtual}} = 47.08\degree$ (59\%), where $\alpha_R=34.09\degree$. The inner angle is $\alpha_{in} = 14.08\degree$. For the given parameters, we expect that the baseline is $b \approx 5\,$cm. Therefore, the minimal distance for recording an average-height person, $h_{\text{avg}} = 1.8\,$m, is $d_{\text{min}}= 3.69$\,m, and the retained common FOV is $\%_{\text{FOV}} = 29.4\%$. The main reason for such a small common FOV are the small mirror sizes.

\section{Calibration and Experiments}
\label{sec:experiments}

We describe and evaluate virtual stereo pair calibration and demonstrate 3D human pose reconstruction.

\subsection{Calibration}

\begin{figure}[h!]
	\centering
	\includegraphics[width=.85\linewidth]{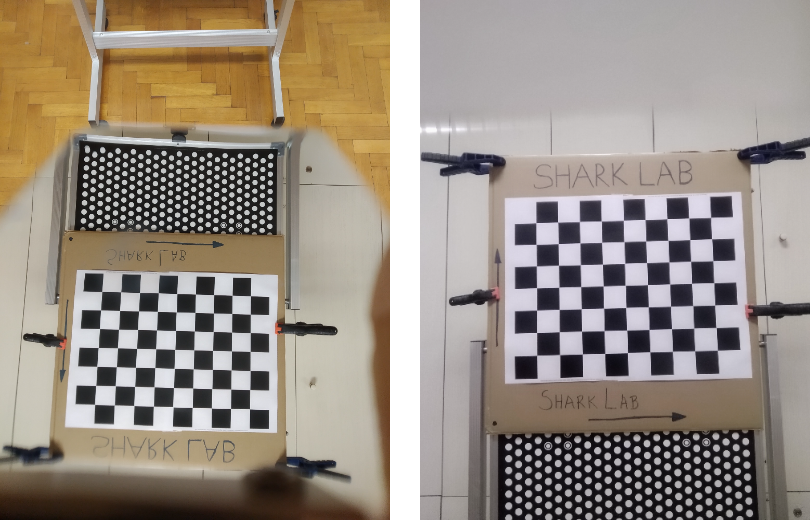}
	\caption{Example of the original calibration chessboard pair of views (before image flipping and rotation), recorded using back and front camera, respectively.}
	\label{fig:calibration-chessboard}
\end{figure}

For stereo pair calibration of two virtual cameras, we use a standard Zhang's calibration \cite{zhang-calibration} in MATLAB. An example of a calibration pair of views is shown in Fig. \ref{fig:calibration-chessboard}. Before the calibration, we flip and rotate the images so that both virtual cameras are upright and horizontally aligned.

\begin{figure}[h!]
	\centering
	\includegraphics[width=.9\linewidth]{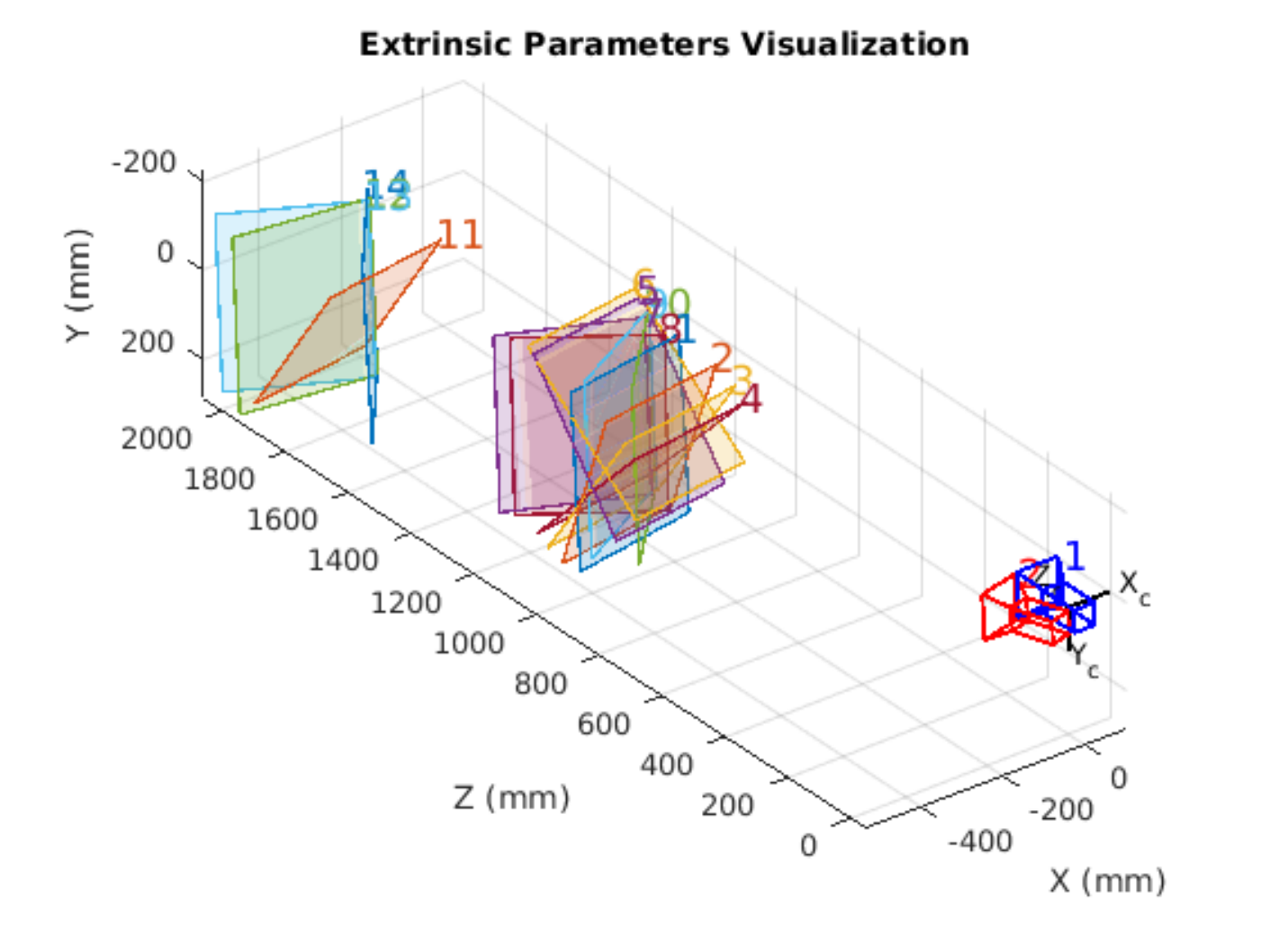}
	\caption{Qualitative evaluation of the calibration. Red and blue cameras represent back and front camera, respectively. The planes on the left represent 14 pairs of views used for the calibration. Most of the views were recorded closer to the calibration chessboard, while others are about 1\,m further.}
	\label{fig:extrinsic-parameters-calibration}
\end{figure}

We record 14 different pairs of views for the calibration. During and after the calibration, the automatic adjustments done by the smartphone, such as autofocus, are turned off. The internal parameters are verified by comparing focal lengths to device's specifications. Qualitative calibration results are shown in Fig. \ref{fig:extrinsic-parameters-calibration}. Regarding extrinsics, there is a slight relative rotation between the mirrors, because the angles are not perfectly set to 45\textdegree{}. The distance between the virtual camera centers is 5.4\,cm, as expected based on the analyses in Sec. \ref{sec:catadioptric-stereo-adapter}.

\begin{table*}[t!]
\centering
\caption{Average distances (mm) between the reconstructed planes and the calibration planes, for each of the 14 views.}
\begin{tabular}{ |c|c|c|c|c|c|c|c|c|c|c|c|c|c|c|c| } 
\hline
Axis/Plane & 1 & 2 & 3 & 4 & 5 & 6 & 7 & 8 & 9 & 10 & 11 & 12 & 13 & 14 & Mean \\
\hline
X & 3.06 & 2.34 & 3.98 & 2.69 & 2.07 & 2.26 & 2.13 & 4.05 & 3.72 & 2.86 & 4.20 & 4.92 & 5.08 & 5.53 & \textbf{3.49} \\ 
Y & 3.11 & 2.37 & 3.98 & 2.72 & 2.07 & 2.27 & 2.13 & 4.07  & 3.72 & 2.86 & 4.21 & 4.96 & 5.11 & 5.54 & \textbf{3.51} \\ 
Z & 4.23 & 3.67 & 4.58 & 2.66 & 1.67 & 2.12 & 1.60 & 3.29 & 4.64 & 3.68 & 4.98 & 4.50 & 5.06 & 6.31 & \textbf{3.79} \\ 
\hline
\end{tabular}
\label{tab:calibration-plane-distances}
\end{table*}

\begin{figure}[h!]
	\centering
	\includegraphics[width=.95\linewidth]{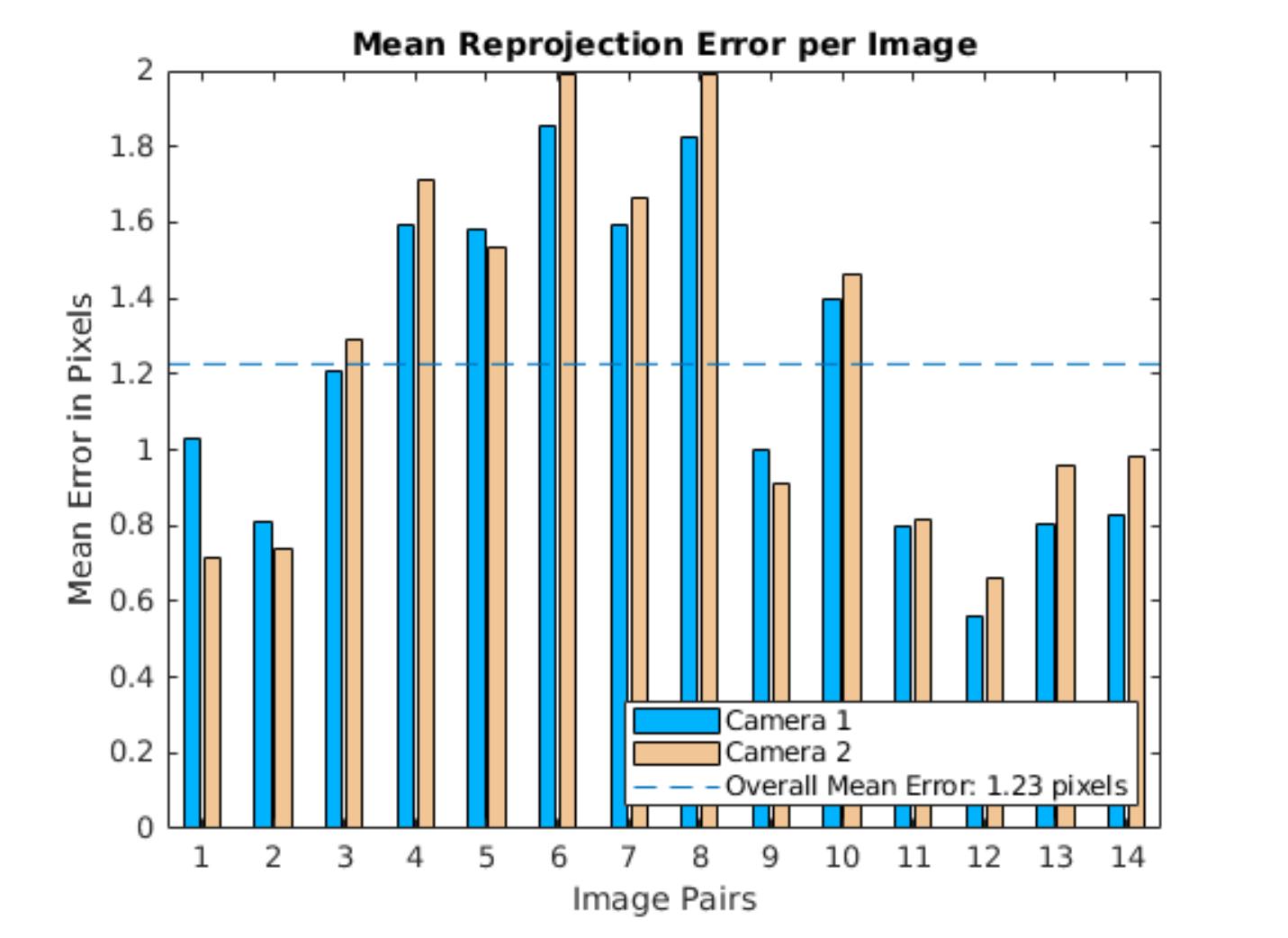}
	\caption{Quantitative evaluation of the calibration. Camera 1 and 2 are back and front camera, respectively.}
	\label{fig:reprojection-errors-calibration}
\end{figure}

For quantitative evaluation, we calculate the distances between the reconstructed points of the calibration plate and the ideal calibration planes (Table \ref{tab:calibration-plane-distances}). Average distances are within few millimeters, which is reasonable. Mean reprojections errors per each image are shown in Fig. \ref{fig:reprojection-errors-calibration}. The overall mean error is 1.23\,pixels. The calibration might be further improved by using more input pairs for the calibration.

\subsection{3D Human Pose Reconstruction}

\begin{figure}[h!]
	\centering
	\includegraphics[width=1.\linewidth]{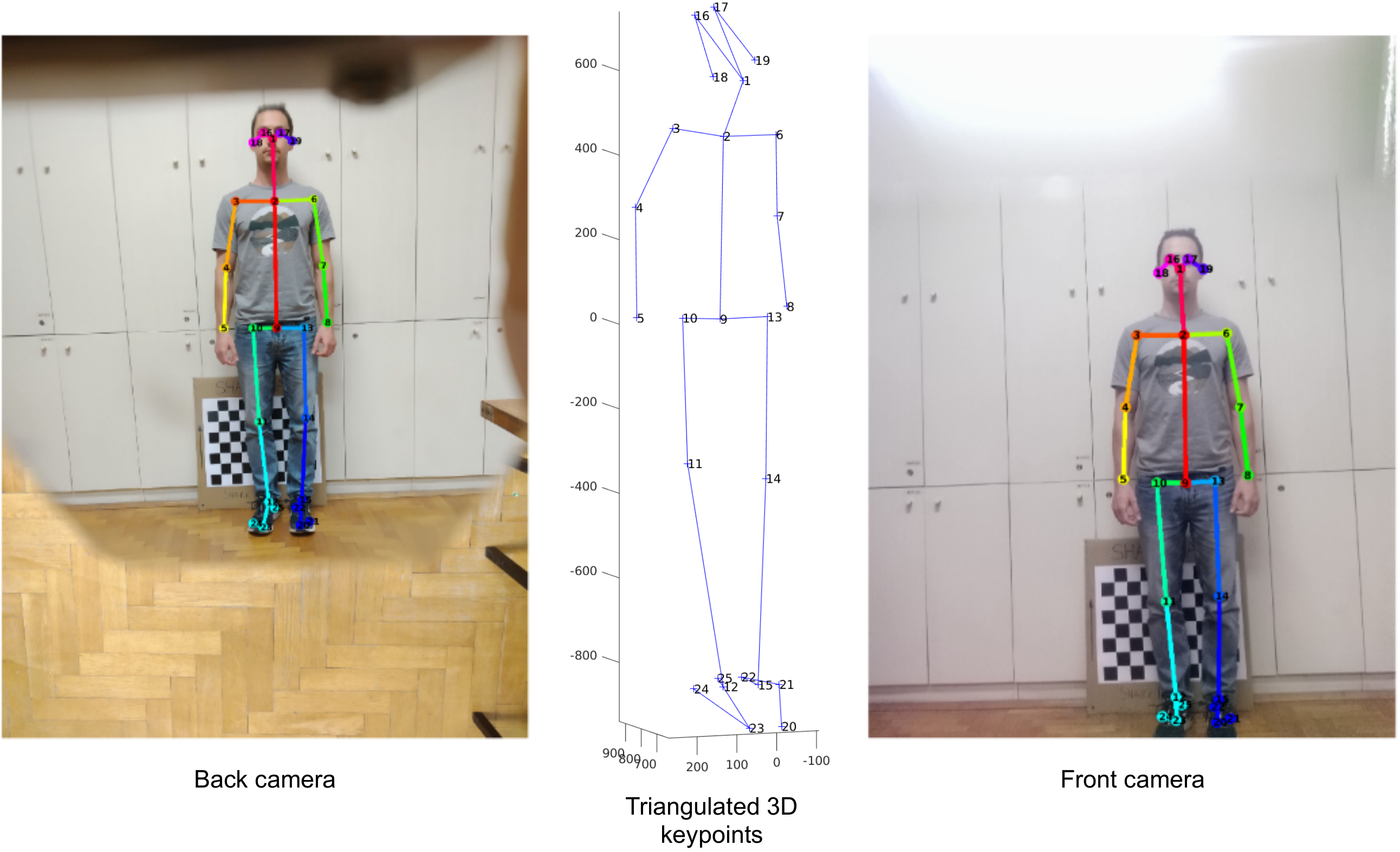}
	\caption{Qualitative reconstruction results. Back and front images, along with the OpenPose keypoints, are shown in the left and right part of the Figure. In the middle, the triangulated 3D skeleton is shown in blue.}
	\label{fig:openpose}
\end{figure}

To demonstrate the working prototype, we reconstruct a 3D human pose. To detect keypoints (human joints), we use OpenPose \cite{openpose}. OpenPose detects 25 corresponding 2D keypoints on both front and back images, as shown in left and right part of Fig. \ref{fig:openpose}. The keypoints are then triangulated to produce the 3D skeleton shown in the middle of Fig. \ref{fig:openpose}.

\begin{table}[h!]
\centering
\caption{Quantitative reconstruction results for the recorded person. Six body measurements are evaluated: lower and upper arm, shoulders and hips width, upper and lower leg, respectively. The first row shows the reconstructed value, the second shows our manual measurement, and the third is the difference between the two. All numbers are shown in cm.}
\begin{tabular}{ |c|c|c|c|c|c|c|c| } 
\hline
 & L.arm & U.arm & Should. & Hips & U.leg & L.leg & Mean \\
\hline
3D & 21.3 & 24.1 & 26.1 & 21.3 & 38.3 & 48.6 & - \\ 
Meas. & 21.2 & 24.7 & 31.0 & 22.1 & 38.9 & 43.5 & - \\ 
\hline
\hline
Diff. & 0.1 & 0.6 & 4.9 & 1.2 & 0.6 & 5.1 & \textbf{2.1} \\ 
\hline
\end{tabular}
\label{tab:body-measurement-evaluation}
\end{table}

To evaluate the reconstruction, we compare several body lengths of the skeleton with our manual measurements: lower arm, upper arm, shoulder width, hips width, upper leg, and lower leg. The measurements, differences and average error are shown in Table \ref{tab:body-measurement-evaluation}. The average error is around 2\,cm, which is an acceptable error for many applications, including anthropometry \cite{anthropometry-review}. Notably, most of the measurements have an error below or around 1\,cm, while shoulder width and lower leg length is much higher. The error in lower leg length might appear due to an error in knee keypoint reconstruction, seen in the middle of Fig. \ref{fig:openpose}. Similar reconstruction failure appears in the right shoulder and elbow, but the error did not propagate to the upper arm segment. 

We attribute most of the reconstruction errors to the fact that the baseline between the virtual cameras is relatively small compared to the distance from which the subject is recorded. Small baseline results in large depth deviations for each correspondent pixel location error.

\section{Conclusion}

The design, analysis, and reconstruction using a prototype catadioptric stereo adapter for front and back smartphone camera is presented. The front-back camera stereo design is compared to previously proposed planar-mirror catadioptric system designs and the analysis of virtual camera FOV based on several degrees-of-freedom is described. The system is successfully calibrated and evaluated based on 3D human pose reconstruction. Taking into account the reconstruction results, we conclude that the reconstruction is successful and the adapter can be used for anthropometric measurements of body lengths. For future work, we propose using wider baseline between the virtual cameras, which requires proportionally larger mirror surfaces, as shown in the analysis. By using a wider baseline, 3D reconstructions should be further improved.

\section*{Acknowledgment}

This work has been supported by the Croatian Science Foundation under the project IP-2018-01-8118. We also thank MSEE Marijan Kuri, Head of Testing Laboratory at the Department of Electronic Systems and Information Processing, for the unconditional help during the configuration of the adapter and mirrors.

\bibliography{references}{}
\bibliographystyle{plain}

\end{document}